\documentclass[twoside,11pt]{article}

\usepackage{jmlr2e}
\usepackage{graphicx}
\usepackage{booktabs}
\usepackage{xcolor}
\usepackage{pifont}
\usepackage{url}
\usepackage{subcaption}
\usepackage{bigfoot}

\newcommand{\cmark}{\ding{51}}
\newcommand{\xmark}{\ding{55}}
\newcommand{\greencheck}{{\color{green}\cmark}}
\newcommand{\redx}{{\color{red}\xmark}}

\ShortHeadings{SimpleDet}{Chen, Han, Li, Huang, Jiang, Wang and Zhang}
\firstpageno{1}

\begin{document}

\title{SimpleDet: A Simple and Versatile Distributed Framework for Object Detection and Instance Recognition}

\author{\name Yuntao Chen \email chenyuntao2016@ia.ac.cn\\
No. 95 Zhongguancun East Road,\\
Haidian District, Beijing 100190, China \\
\AND
\name Chenxia Han \email chenxiahan18@gmail.com
\AND 
\name Yanghao Li \email lyttonhao@gmail.com
\AND
\name Zehao Huang \email zehaohuang18@gmail.com
\AND 
\name Yi Jiang \email jiangyi0425@gmail.com
\AND
\name Naiyan Wang \email winsty@gmail.com
\AND
\name Zhaoxiang Zhang \email zhaoxiang.zhang@ia.ac.cn
}

\editor{}

\maketitle

\begin{abstract}
Object detection and instance recognition play a central role in many AI applications like autonomous driving, video surveillance and medical image analysis. 
However, training object detection models on large scale datasets remains computationally expensive and time consuming. 
This paper presents an efficient and open source object detection framework called \textit{SimpleDet} which enables the training of state-of-the-art detection models on consumer grade hardware at large scale. 
SimpleDet supports up-to-date detection models with best practice. 
SimpleDet also supports distributed training with near-linear scaling out of box. Codes, examples and documents of SimpleDet can be found at \url{https://github.com/tusimple/simpledet}.
\end{abstract}

\begin{keywords}
  Object Detection, Instance Recognition, Distributed Training, Mixed Precision Training
\end{keywords}

\section{Introduction}
\begin{figure}
	\begin{subfigure}{0.32\textwidth}
		\centering
		\includegraphics[scale=0.32]{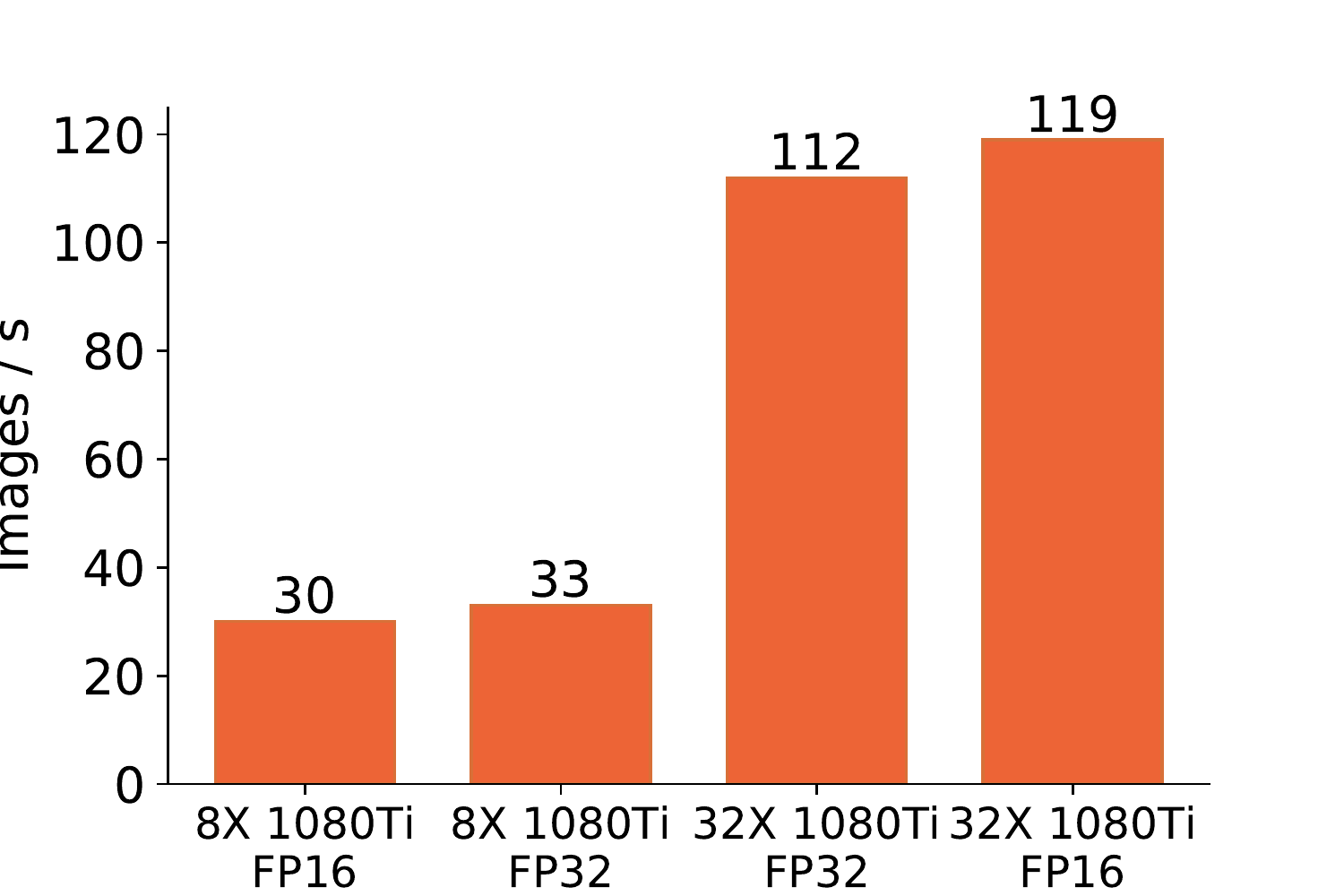}
		\caption{}
		\label{distributed}
	\end{subfigure}
	\begin{subfigure}{0.32\textwidth}
		\centering
		\includegraphics[scale=0.32]{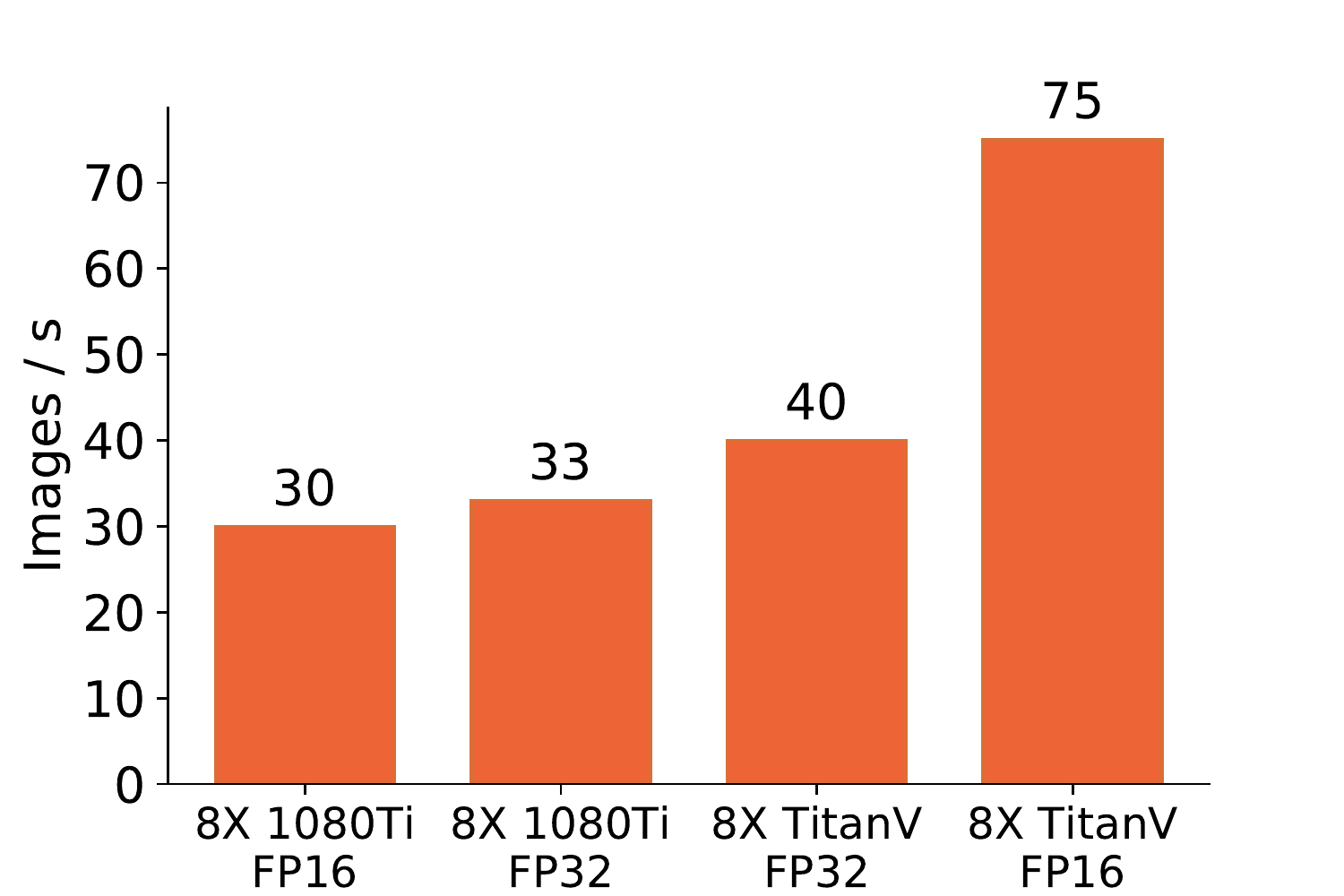}
		\caption{}
		\label{mixed}
	\end{subfigure}
	\begin{subfigure}{0.32\textwidth}
		\centering
		\includegraphics[scale=0.32]{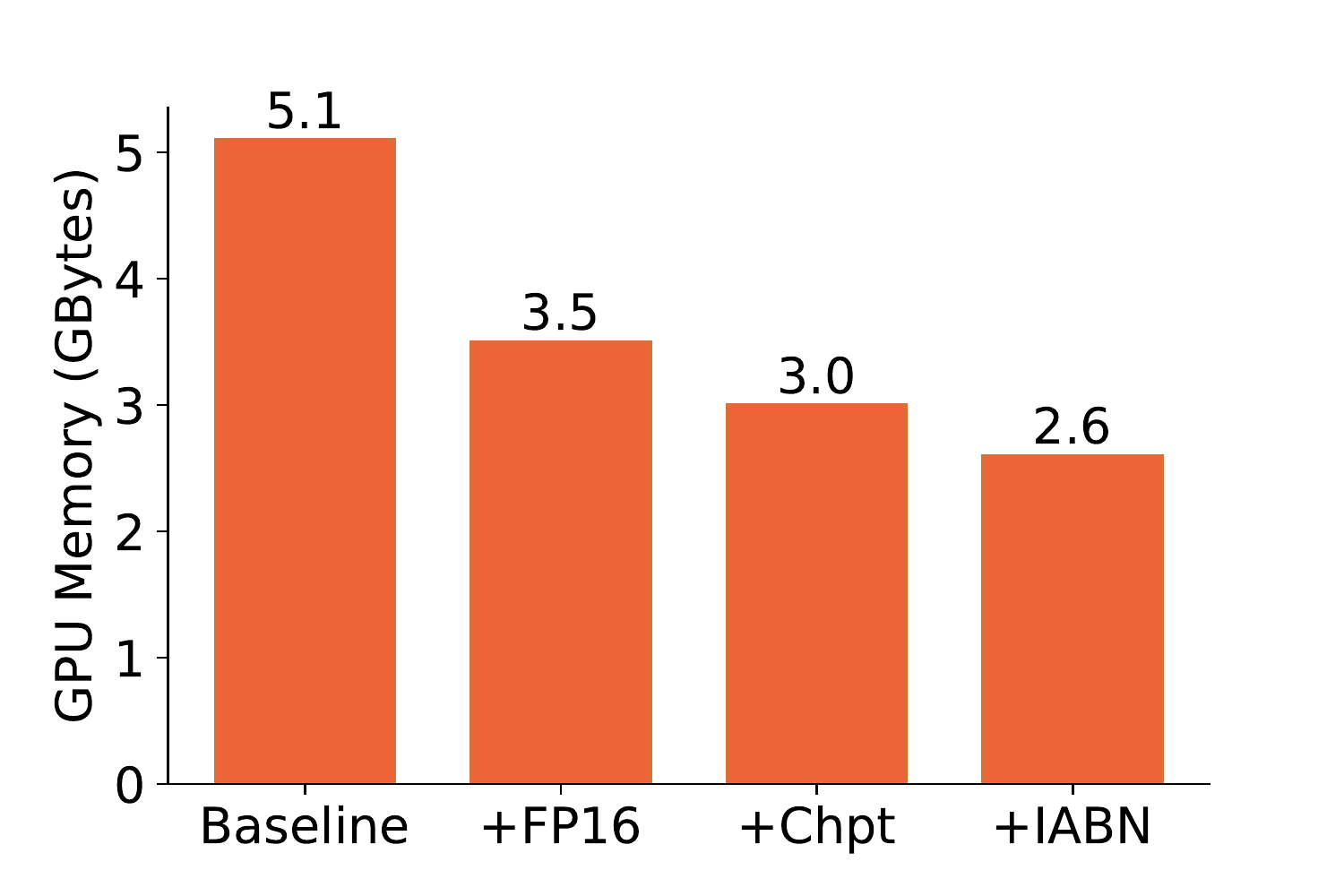}
		\caption{}
		\label{mem}
	\end{subfigure}
	\caption{Results measured on a Faster R-CNN detector with a ResNet-50-C4 backbone. (a) Scaling efficiency of SimpleDet. (b) Speedup of mixed precision training. (c) GPU memory usages of training with mixed precision training(+FP16), memory checkpoint(+Chpt) and in-place activation BN(+IABN)}
\end{figure}

Object detection and instance recognition are at the core of many real-world AI applications like autonomous driving, video surveillance, medical image analysis and cashierless retailing. 
During recent years, more sophisticated detection systems and more challenging datasets have emerged with a ever growing demand in computation power. 
From PASCAL VOC~\citep{voc}($\sim$10K images) to MS COCO~\citep{coco}($\sim$118K images) and to Google Openimages~\citep{kuznetsova2018open}($\sim$1.7M images), the amount of annotated data increases at an incredible speed. 
From AlexNet~\citep{krizhevsky2012imagenet}($\sim$700M FLOPs) to SENet~\citep{hu2018squeeze}($\sim$21G FLOPs), the computation complexity of CNNs also grows beyond imagination. 
These two factors bring the training time of a detection system from several GPU hours to tens of thousands GPU hours, which calls for a distributed detection framework that scales. 
Built on top of MXNet, SimpleDet is the first open source detection framework which provides an efficient batteries-included distributed training system. 
As shown in Figure~\ref{distributed}, our system scales near linearly on a 4-node GPU cluster on the consumer grade 25Gb Ethernet. 
Besides its high efficiency, our system also takes user experience as priority. 
We provide a configuration system in pure python, which eases the use of users and provides great flexibility as the framework and the configuration system are written in the same programming language. 
To further ease the adoption of our system, we provide pre-built Singularity and Docker images. The full codes, examples and documents of SimpleDet can be found at  \url{https://github.com/tusimple/simpledet}. 

\section{Features of SimpleDet}
Like other existing frameworks, SimpleDet covers most state-of-the-art detection models including:
\begin{enumerate}
	\item \verb|Fast RCNN| \citep{girshick2015fast}
	\item \verb|Faster RCNN| \citep{ren2015faster}
	\item \verb|Mask RCNN| \citep{he2017mask}
	\item \verb|Cascade RCNN| \citep{cai2018cascade}
	\item \verb|RetinaNet| \citep{lin2017focal}
	\item \verb|Deformable Convolutional Network (DCN)| \citep{dai2017deformable}
	\item \verb|TridentNet| \citep{li2019scale}
\end{enumerate}
Besides the full coverage of latest models, SimpleDet also provides extensive pre-processing and post-processing routines in detection with their best practices, including various data augmentation techniques, multi-scale training and testing, soft~\citep{bodla2017soft} and weighted NMS, etc. All these features are provided based on the unified and versatile interfaces in SimpleDet, which allows the users to easily customize and extend these features in training.

Apart from these common features, we also would like to highlight several key features of SimpleDet as follows.

\subsection{Distributed Training}

Instead of only scaling up within a single machine, we can utilize data parallel paradigm with more machines to scale out. 
The core of scalable distributed training lies in the efficiency of parameter communication. 
Thanks to the underlying MXNet, SimpleDet supports both parameter server and all-reduce algorithms for model parameter update. 
Along with the mixed precision training technique which will be introduced in the next subsection, SimpleDet could give near-linear scaling efficiency for a 4-node cluster as shown in Figure~\ref{distributed}. 
Note that this performance is only at the cost of consumer grade 25Gb Ethernet in contrast to most of previous works that built on more expensive cross-machine communication hardware, such as InfiniBand. 
We believe this feature could significantly promote the adoption of distributed training.

\subsection{Mixed Precision Training}
Modern specialized hardware like NVIDIA Tensor Core provides 10 times throughput for computation in half precision float (FP16) over single precision one (FP32). 
Besides speed up the training, low precision training also reduces the memory footprint. 
The main obstacle for mixed precision training is the convergence and accuracy drop issue due to the limited range of representation. 
To mitigate this problem, SimpleDet adopts the scale loss proposed by ~\cite{micikevicius2018mixed}. 
In practice, we find that mixed precision training yields identical training curves and detection mAP with full precision training. 

As shown in Figure~\ref{mixed} and ~\ref{mem}, SimpleDet witnesses a 2.0X speedup and a 30\% reduction in memory usage from FP32 training to FP16 training. In addition, mixed precision training effectively reduces the distributed communication cost. This feature also plays an important role in the efficient distributed training.

\subsection{Cross-GPU Batch Normalization}
Due to the limit of GPU memory, modern detectors are trained in a $1 \sim 2$ images per GPU setting. But batch normalization widely used is implemented in a per-GPU manner, which forces researchers to freeze BN parameters during detector training as a workaround. As indicated by ~\cite{peng2018megdet}, freeze-BN detectors trained with linear learning rate scaling scheme fail to converge when the batch size increases beyond a threshold. The failure in convergence harms the scalability of a detection framework. In order to mitigate this limitation, SimpleDet integrates Cross-GPU Batch Normalization(CGBN) and provides a one-line configuration option for users. In practice, we find that scaling a detector to a mini-batch size of 256 with CGBN leads to stable convergence.

\subsection{Memory Saving Technologies}
A limiting factor for the design of new detectors is the amount of memory available for a single GPU. Since the main training paradigm of CNN detector is data parallelism, designs are bound by the amount of memory that a single GPU provides. To mitigate this problem, SimpleDet combines mixed precision training, in-place activation batch normalization~\citep{rota2018place} and layer-wise memory checkpointing~\citep{chen2016training} together to minimize the demand of GPU memory. 
Combining all these techniques, SimpleDet could save up to 50\% memory as in Figure~\ref{mem} with a marginal increase in computation cost compared with the vanilla setting.

\section{Compare with Other Detection Frameworks}
We compare four other detection frameworks with SimpleDet in terms of training speed, supported models and advanced training features in Table~\ref{comparison}.
\begin{enumerate}
	\item \verb|detectron|\footnote{\verb|https://github.com/facebookresearch/Detectron|} is the first general framework for object detection. But its training speed is a major problem as it uses python operators in the core part of the framework extensively. 
	\item \verb|mmdetection|\footnote{\verb|https://github.com/open-mmlab/mmdetection|} is a well-designed framework written in PyTorch which supports a wide range of detection models. Again, the training speed is also a problem for \verb|mmdetection| since it contains many small operations inside the computation graph which incur a large operator invocation overhead.
	\item \verb|tensorpack|\footnote{\verb|https://github.com/tensorpack/tensorpack/tree/master/examples/FasterRCNN|} supports some advanced training features like Cross-GPU Batch Normalization and distributed training, but it lacks supports of some new models. 
	\item \verb|maskrcnn-benchmark|\footnote{\verb|https://github.com/facebookresearch/maskrcnn-benchmark|} is a well optimized framework with amazing training speed. But it supports the least models of all frameworks.
\end{enumerate}

\begin{table}[h]
	\scriptsize
	\centering
	\begin{tabular}{cccccc}
		\toprule
		 & \texttt{detectron} & \texttt{tensorpack} & \texttt{mmdetection} & \texttt{maskrcnn-benchmark} & \texttt{simpledet} \\
		\midrule
		R50-FPN Faster Speed & 29 images/s & 29 images/s & 28 images/s & 40 images/s & 37 images/s \\
		FasterRCNN & \greencheck & \greencheck & \greencheck & \greencheck & \greencheck \\
		MaskRCNN & \greencheck & \greencheck & \greencheck & \greencheck & \greencheck \\
		CascadeRCNN & \greencheck & \greencheck & \greencheck & \redx & \greencheck  \\
		DCN & \redx & \redx & \greencheck & \redx & \greencheck \\
		RetinaNet & \greencheck & \redx & \greencheck & \greencheck & \greencheck \\
		TridentNet & \redx & \redx & \redx & \redx & \greencheck \\
		Cross-GPU BN & \redx & \greencheck & \redx & \redx & \greencheck \\
		Mixed Precision Training & \redx & \redx & \redx & \redx & \greencheck \\
		Distributed Training & \redx & \greencheck & \redx & \redx & \greencheck \\
		Memory Checkpointing & \redx & \redx & \redx & \redx & \greencheck \\
		\bottomrule
	\end{tabular}
	\caption{Comparison of supported features for \texttt{detectron}, \texttt{tensorpack}, \texttt{mmdetection}, \texttt{maskrcnn-benchmark} and \texttt{simpledet} as of Feb 17th, 2019}
	\label{comparison}
\end{table}

\vskip 0.2in
\begin{small}
\bibliography{bib}
\end{small}
\end{document}